\documentclass[11pt, a4paper, logo, copyright, nonumbering]{deepmind}

\usepackage[authoryear, sort&compress, round]{natbib}
\title{Podracer architectures for scalable Reinforcement Learning}

\author[*,1]{Matteo Hessel}
\author[*,1]{Manuel Kroiss}
\author[1]{\\Aidan Clark}
\author[1]{Iurii Kemaev}
\author[1]{John Quan}
\author[1]{Thomas Keck}
\author[1]{Fabio Viola}
\author[*,1]{Hado van Hasselt}

\affil[*]{Equal contributions}
\affil[1]{DeepMind}

\begin{abstract}
Supporting state-of-the-art AI research requires balancing rapid prototyping, ease of use, and quick iteration, with the ability to deploy experiments at a scale traditionally associated with production systems. Deep learning frameworks such as TensorFlow, PyTorch and JAX allow users to transparently make use of accelerators, such as TPUs and GPUs, to offload the more computationally intensive parts of training and inference in modern deep learning systems. Popular training pipelines that use these frameworks for deep learning typically focus on (un-)supervised learning. How to best train reinforcement learning (RL) agents at scale is still an active research area. In this report we argue that TPUs are particularly well suited for training RL agents in a scalable, efficient and reproducible way. Specifically we describe two architectures designed to make the best use of the resources available on a TPU Pod (a special configuration in a Google data center that features multiple TPU devices connected to each other by extremely low latency communication channels).
\end{abstract}

\begin{document}

\maketitle

\section{Introduction}

Reinforcement learning (RL) algorithms have been shown to be capable of performing well on challenging sequential decision problems, ranging from board games \citep{silver2016mastering} to video games \citep{mnih2015} and continuous control \citep{vanHasselt2012,Lillicrap2015}. Many of these advances were powered by the adoption of deep learning \citep{alex2012} in RL.

As in other areas of machine learning, however, the adoption of deep learning comes at non-trivial computational cost. Indeed a steady increase can be observed in the workloads associated to popular RL agents \citep{mnih2015,mnih2016asynchronous,MHSODHPAS18}. This has been driven by two factors: 1) an increase in the number of steps of environment interactions required by agents, due to the tackling of increasingly hard problems, 2) an increase in the amount of compute required, on each step, due to the agent's policy (and other predictions) being parameterized by larger deep neural networks.

This has motivated extensive research on scalable research platforms for (deep) RL \citep{horgan2018distributed,espeholt2018impala,Espeholt2020SEED,Nair15,petrov2018,pmlr-v80-liang18b,stooke2019}. Any such platform must balance enabling easy and rapid prototyping (an essential requirement for research) and optimising execution speed (critical for platforms managing workloads that, until recently, were reserved to production systems rather than research experiments).

In this report, we argue that the compute requirements of large scale reinforcement learning systems are particularly well suited for making use of Cloud TPUs \citep{JouYou17}, and specifically TPU Pods: special configurations in a Google data center that feature multiple TPU devices interconnected by low latency communication channels. We collectively call \texttt{Podracers} the research platforms that we have built to support scalable RL research on TPU Pods, and we describe two such architectures, \texttt{Anakin} and \texttt{Sebulba}, designed to train online agents and decomposed actor-learner agents, respectively. Both have been used in recent deep RL publications \citep{xu2020metagradient,oh2021discovering,zahavy2020selftuning}, and are here presented in full detail for the first time. Both build upon JAX \citep{jax2018github}, to distribute computation and make effective use of TPUs.

\section{Background}
In this section we provide context to understand the systems we describe in the subsequent section.
\vspace{-10pt}

\phantomsection
\subsection{Reinforcement Learning}

Reinforcement learning is a formalism for sequential decision making under uncertainty \citep{sutton2018reinforcement}. Consider a learning system (the \textit{agent}) tasked with learning to act in the world it is embedded in (the \textit{environment}), in order to maximise the total cumulative \textit{rewards} collected while interacting with such an environment. The interaction between agent and environment takes place over a discrete loop: at each step, the environment provides the agent with a (partial) snapshot of its state (an \emph{observation}) together with a scalar feedback (the \emph{reward}), and the agent responds by picking an \emph{action} to execute in the environment. For instance, the learning system may be tasked to learn chess \citep{silver2017}: the agent then selects the moves for one of the players; its environment includes the chess board \emph{and} the opponent; observations encode the state of the board; rewards are $0$ everywhere, except on check-mate, where the agent receives $+1$ for winning or $-1$ for losing.

There are several critical differences between RL and other ML paradigms such as supervised learning \citep{bishop2006,murphy2012}. 1) The data distribution observed by the agent depends on the agent's actions; this means that data is inherently \emph{sequential} and \emph{non-stationary}. 3) It is up to the agent to trade off the exploitation of existing knowledge versus the \emph{exploration} of alternative courses of actions that may appear sub-optimal now, but that might allow the agent to learn even better behaviours. 3) The agent is not told, on each step, what the right action is; rewards give an \emph{evaluation} of how well the agent is performing, but do not indicate what better sequence of actions could have been taken instead. 4) the feedback may be stochastic and delayed; it is up to the agent to correctly \emph{assign credit} \citep{minsky1961steps} for past actions to rewards observed at a later time. 

\vspace{-6pt}
\subsection{A brief overview of JAX}

JAX \citep{jax2018github} is a numerical computing library for machine learning (ML) in Python. Its API for writing numerical functions follows NumPy \citep{harris2020array} closely, but \emph{augments} it with a flexible system for composable program transformation \citep{ward89}, that supports automatic differentiation \citep{nolan1953}, SIMD-programming \citep{flynn1972}, just-in-time compilation \citep{mccarthy1960}, and hardware acceleration via GPUs \citep{barron1973} or TPUs \citep{JouYou17}. 

\emph{Automatic differentiation} is critical to modern ML due to the widespread use of gradient descent \citep{cauchy1847}. JAX supports both forward mode \citep{wengert64} and reverse mode \citep{speelpenning1980} differentiation via the program transformations \texttt{grad}, \texttt{hessian}, \texttt{jacfwd} and \texttt{jacrev} that take a numerical function and return a new function computing the derivative of the original one.

\emph{SIMD programming} is popular in ML, as it is common to apply the same function (e.g. a loss) to a batch of data (e.g. a collection of images and labels). JAX supports SIMD programming via \texttt{vmap} and \texttt{pmap}. The first implements automatic vectorization: given a numerical function, \texttt{vmap} returns a new function that expects inputs with an additional (batch) dimension; this avoids having to manually vectorize your code. Through \texttt{pmap}, JAX also supports large scale data parallelism \citep{Dean2012}, where shards of the same computation can distributed across multiple accelerators or host machines.

\emph{Just-in-time (JIT) compilation} is the compilation of code at run time, rather than prior to execution. The \texttt{jit} primitive in JAX uses XLA \citep{xla2017} to JIT-compile numerical programs. JIT compilation provides substantial speed-ups, and bypasses Python's ``global interpreter lock'' (GIL) \citep{beazley2009}. 

\emph{Hardware acceleration} is achieved in JAX by JIT-compiling code, via XLA, for execution on different hardware (e.g. GPU, TPU).  The user only needs to specify on which device to run each computation.

\begin{figure}[t!]
\begin{center}

\centerline{\includegraphics[trim=0 0.7cm 0 1.3cm,clip,width=\columnwidth]{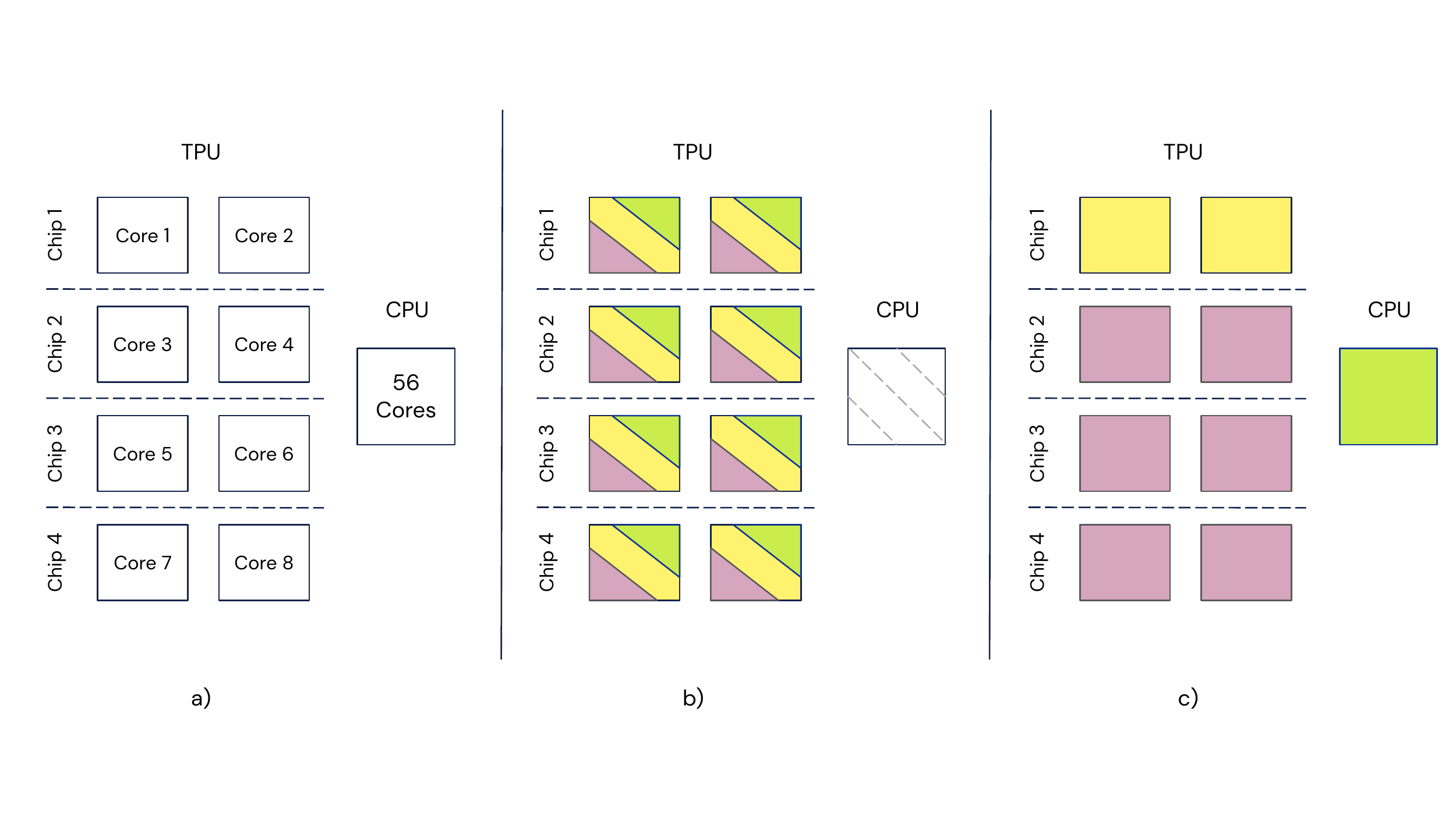}}
\vspace{-20pt}
\caption{\textbf{a)} A graphical depiction of a TPU. One CPU host is connected to 8 TPU cores (grouped into 4 chips). All TPU cores are connected to each other via high speed network. \textbf{b)} In Anakin: environment (in green), action selection (in yellow) and learning (in pink) are all executed on the accelerators, and the computation is replicated across all available cores. \textbf{c)} In Sebulba: the environment computation (in green) happens on the CPU host, the computation associated to action selection (in yellow) and parameter updates (in pink) is distributed across disjoint subsets of the available TPU cores.
}
\vspace{-19pt}
\label{fig:tpus}
\end{center}
\end{figure}

\vspace{-10pt}

\subsection{Tensor Processing Units}

Tensor Processing Units (TPUs) are custom accelerator chips developed by Google to handle machine learning workloads, i.e. workloads associated with big matrix multiplications and batch operations. A TPU pod is a collection of TPU cores that are connected to each other via a high-speed network. Each TPU core (also referred to as a \emph{device}) has access to dedicated high-bandwidth memory (HBM) and can perform computations independently of other cores. TPU cores are managed by CPU machines (henceforth \emph{hosts}) which schedule computations and move data to and from device. On TPUv2 or TPUv3 machines, each host manages 8 TPU cores. A graphical depiction of the host and the 8 associated TPU cores is shown in Figure \ref{fig:tpus}a. 

To run computations on a TPU device, JAX programs must be compiled by XLA on the CPU host, and then loaded onto the TPU device. The computation inputs are transferred from host memory to device memory and the computation program is executed using the device inputs. This produces outputs on the device memory, which can be transferred back to the host when appropriate. There can only be one program running at a time per TPU device. Some computations require the output of the last execution as input to the next execution, in which case the memory may be kept on the device and needless transfers via the host are avoided. Memory on TPU is automatically garbage collected when no Python reference to the corresponding data on the TPU is left.

It is common for data processed on one device to later be needed on a different device. In this case, the data may be transferred directly from a device to another, bypassing the host. Whenever possible, this kind of device to device communication is preferred over host to device communication, since the network transfer speed from a TPU core to another is significantly faster than from CPU to TPU. JAX also offers collective operations (such as \texttt{psum}, to combine tensors and other numerical data across one or more devices). These collective operations are also performed by automatically sending the relevant data directly from a device to another, without requiring any transfer back to the host.

\section{Two Reinforcement Learning Frameworks}
We now describe two frameworks for efficient use of TPUs for reinforcement learning research at scale, called \texttt{Anakin} and \texttt{Sebulba}.  The Anakin framework supports environments that are themselves written in Jax, and hence can run efficiently on the TPU devices.  The Sebulba framework supports arbitrary environments (such as Atari video games) that run on the CPU hosts.

\phantomsection
\subsection{Online Learning with Anakin}

Our first Podracer architecture is \texttt{Anakin}, an online learning system where the environment runs on the TPU alongside the agent. This setup requires the environments themselves to be implemented as JAX (pure) functions, in order for these to be compiled by XLA for execution on the TPU.

Each environment is defined by an \texttt{initial\_state} function to initialize such state and by a \texttt{step} function which maps a state and action to a new state and reward. The state must be handled explicitly for the environment stepping to be a pure function. All the environments must be written in JAX; this restricts the range of environments supported, but provides large benefits in terms of performance and cleanliness of the research platform.

For instance, we found that writing \emph{agents} in this setup is straight forward. There is no need for the actor/learner separation that is so popular in large scale deep RL platforms \citep{horgan2018distributed, espeholt2018impala, Espeholt2020SEED}. Instead, the agent environment interaction loop maps more directly onto the math or pseudo-code you would find in an RL textbook.

\begin{figure}[t!]
\begin{center}

\centerline{\includegraphics[trim=0 0.8cm 0 1.4cm,clip,width=\columnwidth]{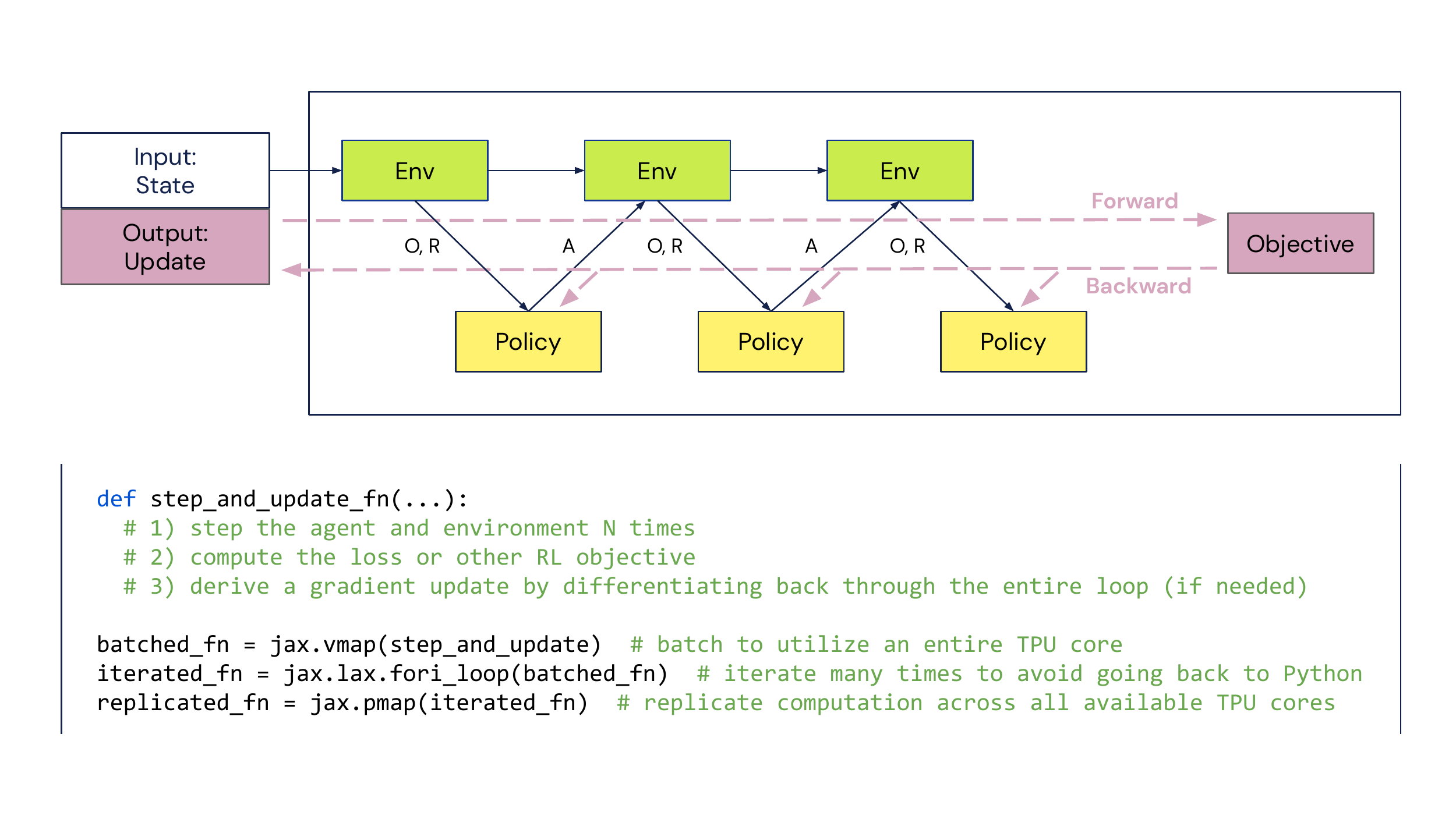}}
\vspace{-5pt}
\caption{\textbf{Top}: The minimal unit of computation in Anakin: a sequence of interactions between one agent and one environment, followed by the calculation of a parameter update (typically optimising for a suitable RL objective -- e.g. a loss). The computation of the update can reuse the computation of the policy, as they are all part of a single JAX function. \textbf{Bottom}: Pseudo-code showing how to scale up such JAX programs through vectorisation and replication across many cores.
}
\vspace{-19pt}
\label{fig:anakin}
\end{center}
\end{figure}

The minimal unit of computation in Anakin is depicted on the top of Figure \ref{fig:anakin}: the environment is stepped as part of the forward computation of a suitable RL objective, and the objective can then be differentiated all the way through multiple agent-environment interaction to update the parameters. The parameter update computation (in pink) can reuse the forward pass of any neural network used by the agent's policy (in yellow), as they are part of the same XLA program. 

Scaling this up is very easy using JAX. The minimal unit of computation described above is first vmapped to vectorise the computation across a batch large enough to ensure good utilisation of an entire TPU core, then the vectorized function is distributed across the 8 cores of a TPU, by using the \texttt{pmap} primitive to replicate the program across all cores.

Figure \ref{fig:tpus}b graphically depicts this: each TPU core includes the computation associated with the environment (in green), action selection (in yellow) and the parameter update (in pink). JAX's collective operations (e.g. \texttt{psum}/\texttt{pmean}) can be used in the update function to ensure that parameter updates are averaged across all participating cores. The CPU host is exclusively used to schedule the computation of this large JAX program but is otherwise unused. We provide a minimal open source demonstration of how Anakin can be implemented in \href{https://colab.research.google.com/drive/1974D-qP17fd5mLxy6QZv-ic4yxlPJp-G?usp=sharing}{this Colab}.

The architecture is highly \emph{efficient} as the agent-environment interaction can be compiled into a single XLA program, allowing the compiler to fuse operations and optimise computation. Furthermore, there are no host to device transfers, as the environment itself lives on the TPU, and there is no overhead from Python, as there is no latency-critical work outside of XLA. This means each TPU core is always fully utilized.

The design is also \emph{scalable}, as it is possible to replicate the basic setup, and therefore use larger TPU slices (potentially, all the 2048 cores of a TPU Pod), with the change of one configuration setting. Finally, Anakin experiments are \emph{self contained} and \emph{deterministic}. Overall, these feature makes this Podracer architecture very flexible, as it can be used for debugging research ideas at a small scale, as well as for running reliable and reproducible experiments at a massive scale.

\begin{figure}[t!]
\begin{center}
\centerline{\includegraphics[trim=0 0.4cm 0 0.1cm,clip,width=\columnwidth]{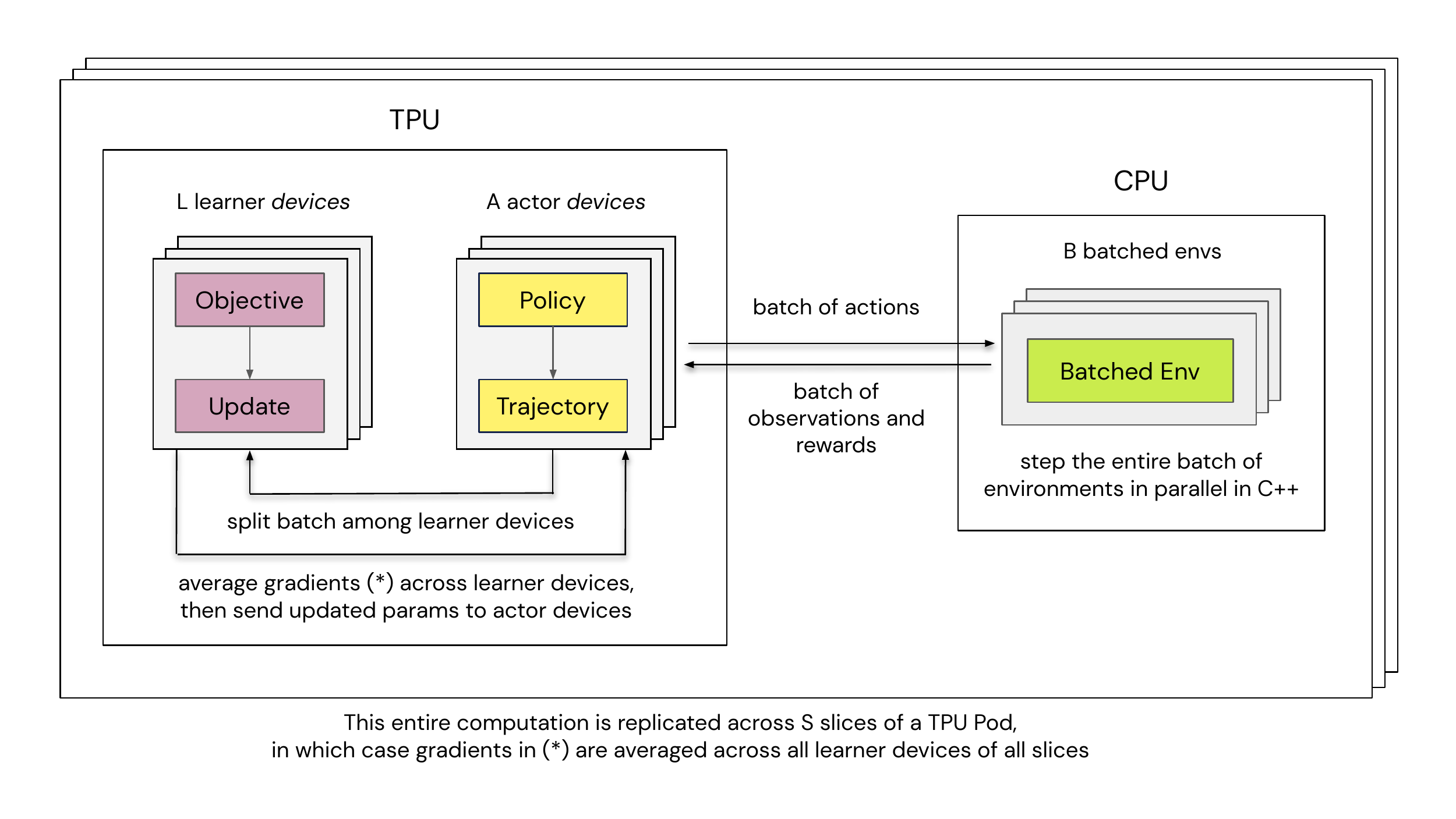}}
\caption{The minimal unit of computation in Sebulba: the 8 TPU cores attached to a CPU host are split between $A$ actor cores and $L=8-A$ learner cores. Actor cores process batches of observations to select actions. Trajectories are accumulated on each actor and then split among all learners, gradient updates are averaged across participating learner cores before updating parameters. Updated parameters are sent directly to the actor devices. The entire computation may be replicated across multiple TPUs.
}
\vspace{-19pt}
\label{fig:sebulba}

\end{center}
\end{figure}

\subsection{Decomposed Actors and Learners with Sebulba}

Our second architecture, Sebulba, relaxes Anakin's assumption that environments can be compiled to run on TPU. Instead, it supports arbitrary environments, and relies on an actor-learner decomposition \citep{espeholt2018impala} for performance. In decomposed architectures, several actors interact with parallel copies of an environment; the experience they generate is fed to a learner through a queue.

As in Anakin, Sebulba co-locates acting and learning on a single TPU machine. However, it steps the environments on the host CPU, and splits the available 8 TPU cores (for each host) in two disjoint sets: $A$ cores are used exclusively to act, and the remaining $8-A$ cores are used to learn. Figure \ref{fig:tpus}c graphically depicts how the computation is distributed across the available hardware: the environment (in green) is executed on the CPU host, the action selection (in yellow) and the parameter update computation (in pink) are assigned to distinct TPU cores. The optimal split between actor and learner cores depends on the nature of the agent being trained. For simple model-free agents we often find it convenient to have 3x as many learner cores as actor cores (since the backward pass is slower than the forward pass). For other agent designs, e.g. if an agent uses search, the optimal split may differ.

To generate experience, we use (at least) one separate Python thread for each actor core. Each Python thread then steps an entire batch of environments in parallel and feeds the resulting batch of observations to a TPU core, to perform inference in batch and select the next batch of actions. To minimise the effect of Python’s GIL, when stepping a batch of environments in parallel, each Python actor-thread interacts with a special batched environment; this is exposed to Python as a single environment that takes a batch of actions and returns a batch of observations; behind the scenes it steps each environment in the batch in parallel using a shared pool of C++ threads. To make efficient use of the actor cores, it is essential that while a Python thread is stepping a batch of environments, the corresponding TPU core is not idle. This is achieved by creating multiple Python threads per actor core, each with its own batched environment. They threads alternate in using the same actor core, without manual synchronization.

To learn from this experience, each actor thread accumulates a batch of trajectories of fixed length on device, splits the batch of trajectories along the batch dimension, sends each shard directly to one of the learners through the fast device-to-device communication channel, and places the Python reference to this tensor data onto a Python queue. A single learner thread on host then takes the handle to the data (already sharded across the appropriate learner cores), and executes the same update function on all the TPU cores dedicated to learning (via JAX's \texttt{pmap} primitive). Each learner core will then apply the same update function to its own shard of the experience, and parameter updates can averaged (by using JAX's \texttt{pmean}/\texttt{psum} primitives) across all participating learner cores.

As for the transfer of trajectories from actor cores to learner cores, the collective operations also exploit the fast device-to-device transfer to avoid transferring parameters back and forth between the host and TPU. If the gradients are averaged before each update, the parameters on the different learner cores remain in sync throughout training without requiring additional transfers. The actor devices receive the new parameters in a timely manner as the learner thread sends them to the actor cores after each update. The Python actor threads can then switch to using the latest parameters before each new inference step.

This minimal unit of computation in Sebulba is graphically depicted in Figure \ref{fig:sebulba}, including the flow of information between the various components. Similar to Anakin, this computation can be scaled up via replication. For every additional replica we gain a separate host CPU (that independently steps its own batched environments and uses the associated actor cores for inference). The learner cores in each replica process only trajectories generated on the corresponding host, but parameter updates will be averaged using JAX's collective operations across all learner cores from all replicas.

\section{Example use cases and performance}

\begin{figure}[t!]
\begin{center}
\centerline{
\includegraphics[width=0.33\columnwidth]{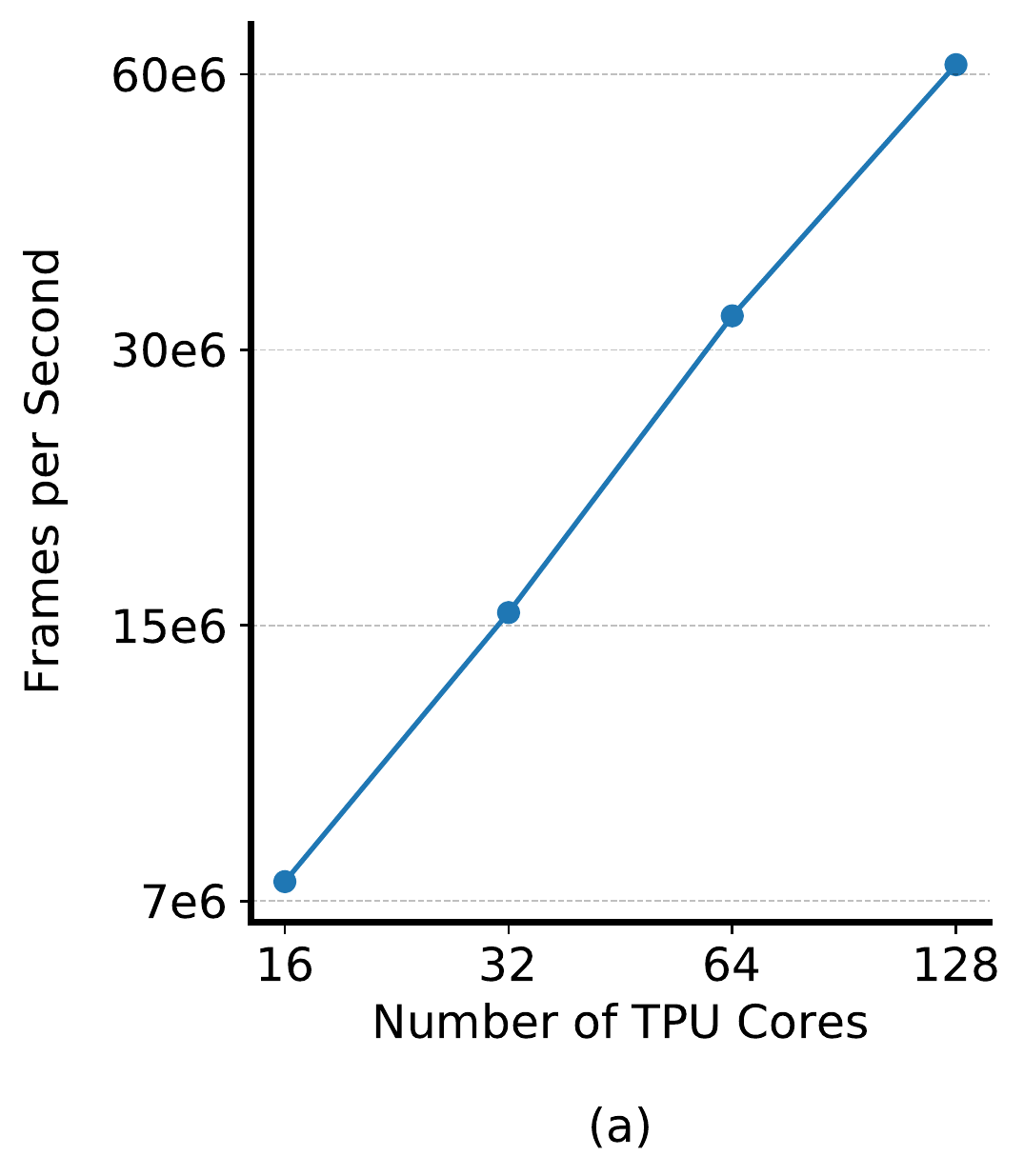}
\includegraphics[width=0.33\columnwidth]{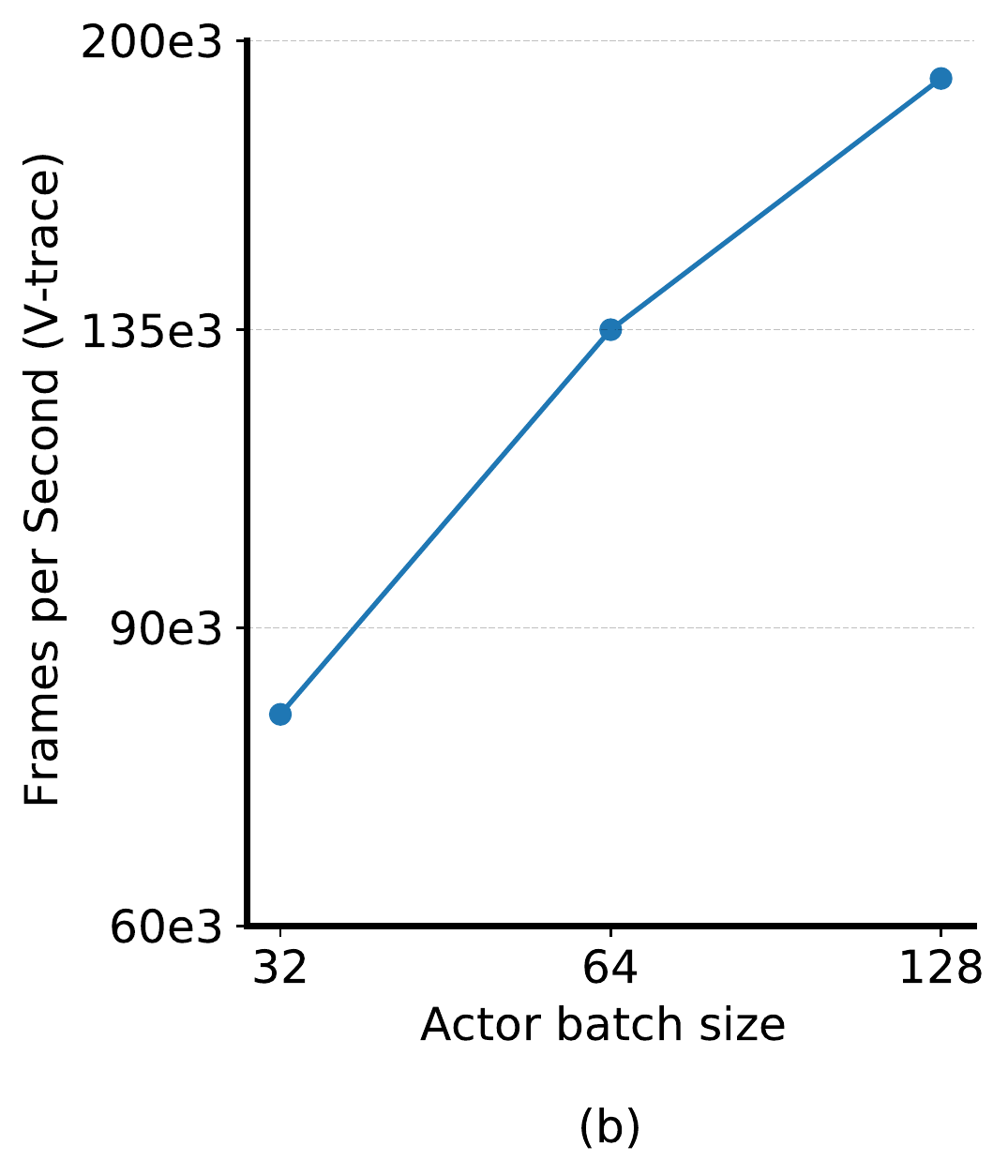}
\includegraphics[width=0.33\columnwidth]{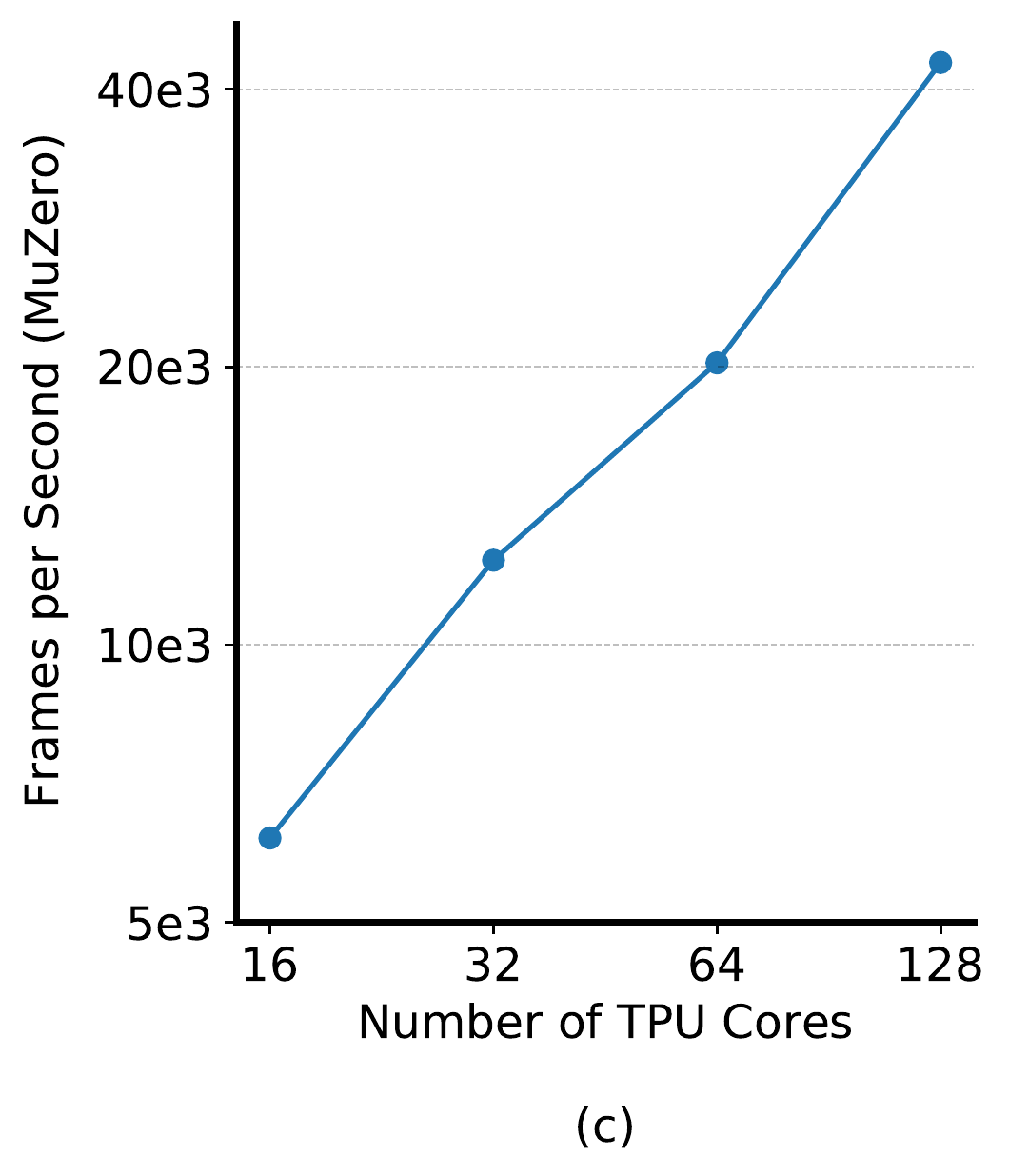}
}
\caption{\textbf{a)} FPS for Anakin, as a function of the number of TPU cores, ranging from 16 (i.e. 2 replicas) to 128 (i.e.16 replicas). \textbf{b)} FPS for a Sebulba implementation of IMPALA's V-trace algorithm, as a function of the actor batch size, from 32 (as in IMPALA) to 128. \textbf{c)} FPS for a Sebulba implementation of MuZero, as a function of the number of TPU cores, from 16 (i.e. 2 replicas) to 128 (i.e. 16 replicas).} 
\label{fig:sebulba_fps}
\end{center}
\end{figure}

In this section we discuss some concrete use cases, to highlight the expected performance (and cost) of using the Anakin and Sebulba architectures in settings that are relevant to today's research.

\subsection{Anakin:}

When using small neural networks and grid-world environments an Anakin architecture can easily perform $5$ million steps per second, even on the 8-core TPU accessible \emph{for free} \footnote{As of April 12 2021, using the hosted TPU runtime.} through Google Colab. This can be very useful to experiment and debug research ideas in the friendly Colab environment.

In Figure \ref{fig:sebulba_fps}a we show how, thanks to the efficient network connecting different TPU cores in a Pod, performance scales almost linearly with the number of cores; the collective operations used to average gradients across replicas appear to cause only minimal overhead.

In a recent paper by \citet{oh2021discovering} Anakin was used, at a much larger scale, to discover a general reinforcement learning update, from experience of interacting with a rich set of environments implemented in JAX. In this paper, Anakin was used to learn a single shared update rule from $60K$ JAX environments and $1K$ policies running and training in parallel.

Despite the complex nature of the system, based on the use of neural networks to meta-learn not just a policy but the entire RL update, Anakin delivered over $3$ million steps per second on a 16-core TPU. Training the update rule to a good level of performance, required running Anakin for approximately 24 hours; this would cost approximately $100$ dollars on GCP's preemptible instances \footnote{As of April 12 2021, using TPU V3, see https://cloud.google.com/tpu/pricing for reference.}.

This use case also demonstrated a potential synergy between Anakin and Sebulba: once a general RL update was learned on JAX environment in Anakin, the meta-learned update rule was lifted and loaded in a Sebulba agent to verify it was general enough to also solve Atari 2600 classic video games.

\subsection{Sebulba:}

Our second podracer architecture has also been extensively used for exploring a variety of RL ideas at scale, on environments that cannot be compiled to run on TPU (e.g. Atari, DmLab and Mujoco).

As both IMPALA and Sebulba are based on a decomposition between actors and learners, agents designed for the IMPALA architecture can be easily mapped onto Sebulba; for instance a Podracer version of the V-trace agent easily reproduced the results from \cite{espeholt2018impala}.

However, we found that training an agent for  $200$ million frames of an Atari game could be done in just $\sim1$ hour, by running Sebulba on a 8-core TPU. This comes at a cost of approximately $2.88$ dollars, on GCP's pre-emptible instances. This is similar in cost to training with the more complex SEED RL framework, and much cheaper than training an agent for $200$ million Atari frames using either IMPALA or single-stream GPU-based system such as that traditionally used by DQN.

Just replicating IMPALA's setup does not however make the best use of the available resources. Sebulba enables to scale up in different ways which are not exploited by the canonical IMPALA: using bigger batch size in the actors, increasing the batch size on the learner (by using longer trajectories and/or more replicas), and using larger or deeper neural networks.

Increasing the batch size is particularly effective at maximising throughput. In Figure \ref{fig:sebulba_fps}b, we report the effect of the actor batch size when training a V-trace agent on an 8-core TPU, with a trajectory length of 60 (up from 20 in IMPALA) and varying actor batch size. We found increasing the batch size from 32 (as in IMPALA) to 128 allowed the same V-trace agent to reach $200K$ frames per second on an 8 core TPU.

In addition to the trajectory length the effective batch size used to compute each update also depends on how many times we replicate the basic 8-TPU setup. Sebulba also scales effectively along this dimension: using 2048 TPU cores (an entire Pod) we were able to further scale all the way to $43$ million frames per second, solving the classic Atari videogame \texttt{pong} in less than 1 minute.

One disadvantage of maximising throughput by increasing the batch size on actors and learners is that, if done naively, it can result in much reduced data efficiency. When interested in maximising data efficiency instead of wall-clock time, we found it convenient to scale in a different way: maximising TPU utilisation by using larger networks instead of bigger batches.

The Sebulba architecture was quite effective in this regard as well, by increasing the number of channels and number of residual blocks in the IMPALA network, we found we could improve significantly the data efficiency of our agents. Since this is achieved by improving TPU utilisation rather than increasing the total number of TPU hours, it does not increase the cost of the experiments.

Sebulba has also been used to train search-based agents inspired by MuZero \cite{schrittwieser2019}. The workloads associated to these agents are very different from that of model-free agents like IMPALA. The key difference is in the cost of action selection. This increases because MuZero's policy combines search with deep neural networks (used to guide and/or truncate the search).

Typically, search-based agents like MuZero required custom C++ implementations of the search to deliver good performance. We could reproduce results from MuZero (no Reanalyse) on multiple RL environments, using Sebulba and a pure JAX implementation of MCTS. Training a MuZero agent with Sebulba for $200M$ Atari frames takes $9$ hours on a 16-core TPU (at a cost of $\sim40$ \$ on GCP's preemptible instances).

We found that scalability, via replication, was particularly useful in this context. Figure \ref{fig:sebulba_fps}c reports the number of frames per seconds processed by Sebulba when running MuZero on Atari, as a function of the number of TPU cores. The throughput increased linearly with the number of cores.

Since in this setting the search was the main bottleck we found that we could efficiently decouple the acting batch size from the learning batch size, by splitting each batch on the learner(s), and then applying N updates instead of a single larger one. This simple trick allowed to preserve the same degree of data efficiency while scaling up; therefore the linear speed up provided by larger Pod slices delivered faster turnaround time for research on search based agents, without increasing the cost of training an agent to any given level of performance.


\section{Conclusions:}
We find that the Podracer architectures described in this report, and implemented in JAX, are very effective in supporting a variety of research use cases; they find a sweet spot between scalability, on one side, and ease of implementation and maintainance on the other. Similarly to \citet{Espeholt2020SEED}, we also find that RL platforms based on TPUs can deliver exceptional performance at a cost that is often better to that of training smaller scale agents that lack parallelism or TPU acceleration.

\section{Acknowledgements:}
Thanks to all our great colleagues at DeepMind, and in particular to Trevor Cai, Dan Horgan, Hamza Merzic, Dan Belov for all the great comments, discussions and contributions. Also, thanks the JAX team for the amazing JAX library that made implementing all these architectures easy and fun!

\bibliography{main}

\end{document}